\documentclass[conference]{IEEEtran}
\IEEEoverridecommandlockouts

\usepackage{cite}
\usepackage{amsmath,amssymb,amsfonts}
\usepackage{hyperref} 

\usepackage{algorithmic}
\usepackage{booktabs}
\usepackage{graphicx}
\usepackage{textcomp}
\usepackage{xcolor}
\usepackage{tabularray}
\usepackage[many]{tcolorbox}

\def\BibTeX{{\rm B\kern-.05em{\sc i\kern-.025em b}\kern-.08em
    T\kern-.1667em\lower.7ex\hbox{E}\kern-.125emX}}
    
\begin{document}

\title{Are Large Language Models Capable of \\ Deep Relational Reasoning? Insights from DeepSeek-R1 and Benchmark Comparisons\\

\thanks{This work is supported by the Research Grants Council of the Hong Kong Special Administrative Region, China under Grant UGC/FDS24/H25/24.}
}

\author{\IEEEauthorblockN{Chi Chiu So}
\IEEEauthorblockA{\textit{School of Professional Education and Executive Development} \\
\textit{The Hong Kong Polytechnic University}\\
Hong Kong, China \\
kelvin.so@cpce-polyu.edu.hk}
\and
\IEEEauthorblockN{Yueyue Sun\IEEEauthorrefmark{1}, Jun-Min Wang\IEEEauthorrefmark{2}}
\IEEEauthorblockA{\textit{School of Mathematics and Statistics} \\
\textit{Beijing Institute of Technology}\\
Beijing, China \\
\IEEEauthorrefmark{1}yysunnya@bit.edu.cn, \IEEEauthorrefmark{2}jmwang@bit.edu.cn}
\and
\IEEEauthorblockN{Siu Pang Yung}
\IEEEauthorblockA{\textit{Department of Mathematics} \\
\textit{The University of Hong Kong}\\
Hong Kong, China \\
spyung@hku.hk}
\and
\IEEEauthorblockN{Anthony Wai Keung Loh\IEEEauthorrefmark{3}, Chun Pong Chau\IEEEauthorrefmark{4}}
\IEEEauthorblockA{\textit{School of Professional Education and Executive Development} \\
\textit{The Hong Kong Polytechnic University}\\
Hong Kong, China \\
\IEEEauthorrefmark{3}anthony.wk.loh@cpce-polyu.edu.hk, \IEEEauthorrefmark{4}james.chau@cpce-polyu.edu.hk}
}

\maketitle

\begin{abstract}
How far are Large Language Models (LLMs) in performing deep relational reasoning? In this paper, we evaluate and compare the reasoning capabilities of three cutting-edge LLMs, namely, DeepSeek-R1, DeepSeek-V3 and GPT-4o, through a suite of carefully designed benchmark tasks in family tree and general graph reasoning. Our experiments reveal that DeepSeek-R1 consistently achieves the highest F1-scores across multiple tasks and problem sizes, demonstrating strong aptitude in logical deduction and relational inference. However, all evaluated models, including DeepSeek-R1, struggle significantly as problem complexity increases, largely due to token length limitations and incomplete output structures. A detailed analysis of DeepSeek-R1's long Chain-of-Thought responses uncovers its unique planning and verification strategies, but also highlights instances of incoherent or incomplete reasoning, calling attention to the need for deeper scrutiny into LLMs' internal inference dynamics. 
We further discuss key directions for future work, including the role of multimodal reasoning and the systematic examination of reasoning failures. Our findings provide both empirical insights and theoretical implications for advancing LLMs' reasoning abilities, particularly in tasks that demand structured, multi-step logical inference. 
Our code repository will be publicly available at \href{https://github.com/kelvinhkcs/Deep-Relational-Reasoning}{https://github.com/kelvinhkcs/Deep-Relational-Reasoning}.
\end{abstract}


\section{Introduction}

Large Language Models (LLMs) have demonstrated remarkable capabilities across a broad spectrum of tasks, such as time-series analysis~\cite{zhou2024can,jin2023time} and AI-driven education~\cite{wen2024ai,kirova2024software}. In particular, the recent release of DeepSeek-R1~\cite{guo2025deepseek}, the first LLM equipped with deep reasoning capabilities, has spurred growing optimism that LLMs can now perform complex logical reasoning at a level potentially surpassing human proficiency.

With the rise of LLMs, increasing attention has been directed toward understanding their reasoning abilities, widely considered a critical bottleneck in the pursuit of Artificial General Intelligence (AGI)~\cite{feng2024far,mohammad2023llm}. Prior studies have primarily evaluated reasoning performance through benchmarks involving arithmetic reasoning~\cite{yan2024survey,fang2024mathodyssey} and symbolic reasoning~\cite{gui2024logicgame,xiao2024logicvista} etc. While many LLMs developed in recent years~\cite{yuan2023scaling,zong2023solving} have achieved impressive results on these benchmarks, they largely lack explicit deep reasoning mechanisms. Preliminary work has begun to explore the reasoning abilities of more recent LLMs with deep reasoning capabilities, such as DeepSeek-R1, in some reasoning domains, for example, arithmetic reasoning~\cite{alonso2025mathematics}. However, a comprehensive understanding of their general logical reasoning capacity remains underexplored.

Concurrently, there has been ongoing debate about whether the existing reasoning benchmarks are truly indicative of deep reasoning ability. Critics argue that many of these benchmarks require only shallow or pattern-based reasoning~\cite{valmeekam2023planbench}, making them insufficient for validating the sophisticated reasoning capabilities of modern LLMs. One key area that warrants deeper investigation is relational reasoning, which is the ability to understand and reason about relationships between entities. This faculty is a cornerstone of intelligence, closely tied to logical thinking and problem-solving in unfamiliar contexts~\cite{crone2009neurocognitive,halford1998processing}. Recent research has also suggested that relational reasoning may remain a fundamental weakness of current LLMs~\cite{li2024llms}.

Motivated by these gaps, this paper investigates the relational reasoning abilities of LLMs with deep reasoning capabilities, particularly DeepSeek-R1, by comparing their performance against other widely-used LLMs. Our study aims to:

\begin{itemize}
    \item Evaluate the relational reasoning performance of state-of-the-art LLMs with deep reasoning capabilities, with a focus on DeepSeek-R1;

    \item Explore the limitations of these models in complex relational problems, particularly in scenarios that remain underexamined in existing literature;

    \item Derive actionable insights to inform future LLM development and research directions.
\end{itemize}

To conduct a thorough evaluation, we propose two classes of relational reasoning benchmarks: family tree and general graph reasoning, using a zero-shot prompting setup. Our family tree benchmark involves identifying four distinct family tree relations - sister, aunt, grandson, and paternal great aunt - within large and complex family structures, while our general graph reasoning benchmark include node connectivity and shortest-path problems.

Our experimental findings reveal several key insights:

\begin{enumerate}
    \item  LLMs with deep reasoning capabilities, such as DeepSeek-R1, significantly outperform prior models without such capabilities in deep relational reasoning problems.

    \item The primary bottleneck in these models lies in the token limit inherent to current LLM architectures, which restricts their capacity to handle large relational structures.

    \item While DeepSeek-R1 demonstrates strong planning, verification, and step-by-step deduction abilities, some intermediate steps are unstructured, suggesting that, despite appearing coherent, the underlying reasoning rationale may not always be fully sound.
\end{enumerate}

This paper offers the following key contributions:

\begin{enumerate}
  \item To the best of our knowledge, this is the first comprehensive evaluation of relational reasoning in state-of-the-art LLMs with deep reasoning capabilities, with a particular focus on DeepSeek-R1.

  \item We introduce and apply an experimental pipeline of two representative benchmarks, family tree and general graph reasoning, designed to systematically evaluate LLMs' relational reasoning capacity in a zero-shot prompting manner.

  \item We demonstrate that DeepSeek-R1, known for its long Chain-of-Thought reasoning enabled by reinforcement learning, fundamentally outperforms other LLMs (e.g., GPT-4o and DeepSeek-V3) in relational reasoning problems.
\end{enumerate}

The rest of the paper is organized as follows: Section~\ref{sec:2} reviews related work. Section~\ref{sec:3} introduces our two benchmarks: family tree and general graph reasoning. Section~\ref{sec:4} details our experimental setup, including the models under evaluation, the data generation processes, and the procedure and metrics of evaluation. Section~\ref{sec:5} presents our experimental results and an in-depth discussion. Section~\ref{sec:6} concludes the paper. Our code repository will be publicly available at \href{https://github.com/kelvinhkcs/Deep-Relational-Reasoning}{https://github.com/kelvinhkcs/Deep-Relational-Reasoning}.

\section{Related Works}\label{sec:2}

\subsection{Large Language Models}

\paragraph{Large Language Models} 
Large Language Models (LLMs) refer to generative, pre-trained, Transformer-based AI models. By leveraging massive corpora of data representing human knowledge, along with advanced pre-training and finetuning techniques, LLMs have demonstrated unprecedented, human-like reasoning capabilities and have achieved state-of-the-art performance in problems once considered infeasible, such as machine translation~\cite{hendy2023good}, music composition~\cite{yuan2024chatmusician}, and more. These models are typically trained using an autoregressive loss function to maximize the likelihood of predicting the next token given the preceding context~\cite{vaswani2017attention}.

\paragraph{Multimodal LLMs} 
Unlike standard LLMs that process only textual input, multimodal LLMs are capable of understanding and generating responses from diverse input types, such as images, audio, and video~\cite{lee2023multimodal, wu2024next}. A well-known example is Contrastive Language–Image Pre-training (CLIP)~\cite{radford2021learning}. These models typically employ a unified embedding mechanism that enables the joint processing of multiple modalities by embedding and concatenating their token representations into a single sequence.

\paragraph{Logical Reasoning of LLMs} 
With the rapid evolution of LLMs, an increasing body of research has focused on their logical reasoning abilities, particularly in domains such as arithmetic reasoning~\cite{yan2024survey,fang2024mathodyssey} and symbolic reasoning~\cite{gui2024logicgame,xiao2024logicvista}. While earlier LLMs have demonstrated strong performance on such benchmarks, recent studies reveal that models lacking explicit deep reasoning capabilities still struggle with problems requiring structured decision-making, such as the ``blocksworld'' problem, which involves arranging blocks in a specific order~\cite{valmeekam2023planbench}. Moreover, despite the emergence of more powerful models with purported deep reasoning abilities, comprehensive evaluation of their performance on general logical reasoning problems remains limited and urgently needed.

\subsection{Prompting Techniques}

\paragraph{Few-shot Prompting} 
Few-shot prompting involves providing LLMs with a few guiding examples within the prompt to help the model better understand the task and generate more coherent and structured outputs~\cite{brown2020language, wei2022chain}. This foundational strategy has inspired several prompt engineering variants, including Chain-of-Thought (CoT) prompting~\cite{wei2022chain}, Chain-of-Verification (CoVe) prompting~\cite{dhuliawala2023chain}, and Tree-of-Thought (ToT) prompting~\cite{yao2023tree}. These approaches aim to introduce intermediate reasoning steps or verification structures that guide the model through more rigorous logical sequences.

\paragraph{Zero-shot Prompting} 
Zero-shot prompting refers to querying LLMs without providing any explicit problems examples~\cite{kojima2022large,yong2023prompt}. This approach relies solely on the model’s pre-trained knowledge and reasoning capabilities, offering flexibility in unseen problems. However, it often comes with trade-offs in performance and robustness, especially for problems requiring structured, multi-step reasoning.

\paragraph{Multimodal Prompting} 
Multimodal prompting involves querying multimodal LLMs with inputs spanning multiple modalities - text, images, audio, or video~\cite{wasim2023vita}. This technique enables the model to engage in more holistic forms of reasoning, such as spatial, visual, and geometric reasoning. Recent studies suggest that multimodal prompting enhances reasoning robustness and generalization by mimicking the diverse ways in which humans process and interpret information~\cite{ge2024worldgpt,wu2024next}.

\section{Relational Reasoning Benchmarks}\label{sec:3}

In this section, we introduce the two relational reasoning benchmarks used in our study: family tree and general graph reasoning. These benchmarks are widely adopted in relational reasoning literature [32]–[36], especially for evaluating the capabilities of neural network-based models~\cite{dongneural,zimmer2023differentiable,evans2018learning,manhaeve2021neural}.

\subsection{Family Tree Reasoning} \label{sec3:family_tree}

The family tree reasoning benchmark consists of problems that require the deduction of complex family relationships based on a set of basic relations. Specifically, a family tree is constructed using four fundamental predicates:

\begin{itemize}
    \item  \texttt{IsMother}$(x, y)$: True if $y$ is the mother of $x$.

    \item  \texttt{IsFather}$(x, y)$: True if $y$ is the father of $x$.

    \item  \texttt{IsSon}$(x, y)$: True if $y$ is the son of $x$.

    \item  \texttt{IsDaughter}$(x, y)$: True if $y$ is the daughter of $x$.
\end{itemize}

These basic relations are used to infer four higher-order target relations as described below:

\begin{enumerate}
    \item 
    \texttt{HasSister}$(x)$:
    Determines whether $x$ has a sister.
    Formally, this is evaluated as:
    $\exists \: y \not = x, \exists \: z$ such that $\texttt{IsDaughter}(z, y) \wedge \texttt{IsDaughter}(z, x)$.
    
    \item \texttt{IsGrandson}$(x, y)$:
    Determines whether $y$ is the grandson of $x$.
    Formally, this is evaluated as:
    $\exists \: z$ such that $\texttt{IsSon}(z, y) \wedge (\texttt{IsSon}(x, z) \vee \texttt{IsDaughter}(x, z))$.
    
    \item 
    \texttt{IsAunt}$(x, y)$:
    Determines whether $y$ is the aunt of $x$.
    Formally, this is evaluated as:
    $\exists \: z$ such that $\texttt{IsDaughter}(z, y) \wedge \texttt{IsGrandparent}(x, z) \wedge \neg \texttt{IsMother}(x, y)$,
    where $\texttt{IsGrandparent}(x, z)$ is evaluated by checking if $\exists \: w$ such that
    $((\texttt{IsSon}(z, w) \wedge \texttt{IsFather}(x, w)) \vee (\texttt{IsDaughter}(z, w) \wedge \texttt{IsMother}(x, w)))$.
    
    \item 
    \texttt{IsPaternalGreatAunt}$(x, y)$:
    Determines whether $y$ is the paternal great aunt of $x$.
    Formally, this is evaluated as:
    $\exists \: z, w$ such that $\texttt{IsGrandfather}(x, z) \wedge \texttt{IsParent}(z, w) \wedge \texttt{IsDaughter}(w, y)$,
    where $\texttt{IsGrandfather}(x, z)$ holds if $\exists \; v$ such that $\texttt{IsFather}(v, z) \wedge (\texttt{IsSon}(v, x) \vee \texttt{IsDaughter}(v, x))$.

\end{enumerate}

Other family relations, for example, \texttt{IsUncle}$(x, y)$ and \texttt{IsGrandparent}$(x, y)$), can be derived similarly. However, due to logical symmetry and comparable reasoning depth, we adopt the four relations above as representative problems for our family tree reasoning benchmark.

\subsection{General Graph Reasoning} \label{sec:graph_reasoning}

The general graph reasoning benchmark involves problems that require inferring high-level structural properties of graphs from a basic relation:

\begin{itemize}
    \item \texttt{HasEdge}$(x, y)$: True if there is a directed edge from node $x$ to node $y$.
\end{itemize}

From this base relation, we define two key target problems:

\begin{enumerate}
    \item 
    \texttt{Connectivity}$(x, y)$:
    Determines whether a path exists from node $x$ to node $y$.

    \item
    \texttt{Shortest}$(x, y)$:
    Determines the number of edges on the shortest path from node $x$ to node $y$.
\end{enumerate}

While both problems can be solved using classical graph algorithms, for example, Breadth-First Search (BFS) and Dijkstra’s algorithms, our objective is not to test algorithmic knowledge. Instead, we prompt LLMs with natural language descriptions (see Section~\ref{sec4:setup})) to assess their relational reasoning abilities, rather than their ability to mimic algorithmic behavior.

Other simpler problems in the general graph reasoning domain include:

\begin{itemize}
    \item 
    \texttt{4-Connectivity}$(x, y)$: Determines if there is a path from node $x$ to node $y$ within four edges.

    \item 
    \texttt{2-Outdegree}$(x)$: Determines if node $x$ has exactly two outgoing edges.
\end{itemize}

These problems are relatively straightforward, as they only require local neighborhood examination and do not challenge the model’s ability to conduct deep, multi-step reasoning across the entire graph.

Therefore, we focus exclusively on the problems of \texttt{Connectivity}$(x, y)$ and \texttt{Shortest}$(x, y)$ as they necessitate a more sophisticated, global understanding of the graph structure, mirroring the complexity found in our family tree reasoning benchmark problems. This allows us to conduct a more comprehensive and meaningful evaluation of the relational reasoning capabilities of LLMs.

\section{Experiments}\label{sec:4}

This section outlines the LLMs evaluated in our study, followed by our data generation process, our experimental setup and the evaluation metrics adopted.

\subsection{Models}

To comprehensively assess deep relational reasoning capabilities, we select DeepSeek-R1 as our primary model due to its demonstrated strengths in complex reasoning and its accessibility for large-scale testing. For meaningful comparative analysis, we also include DeepSeek-V3 and GPT-4o to represent models without deep reasoning capabilities.

\subsubsection{DeepSeek-R1 (671B parameters)}
Released in early 2025, DeepSeek-R1 is a state-of-the-art LLM comprising 671 billion parameters. It is widely recognized for its remarkable reasoning performance and cost-efficiency, achieved through large-scale reinforcement learning. Notably, it is the first LLM to integrate several novel techniques, including Mixture of Experts (MoE) and Group Relative Policy Optimization (GRPO), to simultaneously improve reasoning and computational efficiency. The model is specifically trained to construct long Chain-of-Thought (CoT) reasoning paths, which are vital for solving problems requiring complex logical deduction across extended contexts~\cite{guo2025deepseek,liu2025logical}.

\subsubsection{DeepSeek-V3}
DeepSeek-V3 is a smaller and more lightweight version of DeepSeek-R1. While it omits full implementations of MoE and long CoT frameworks, it retains core advancements in reinforcement learning and structured reasoning. It serves as a representative model without full deep reasoning capabilities, allowing for fair performance comparisons.

\subsubsection{GPT-4o}
GPT-4o, a leading model in the GPT series, offers competitive reasoning capabilities and long-context understanding. However, it does not adopt the explicit long CoT methodology employed in DeepSeek-R1. It is included as a strong benchmark model without a dedicated deep reasoning architecture.

\subsection{Data Generation}\label{sec4:setup}

The data generation process consists of three primary steps: 1) Adjacency and Boolean Matrix Construction, 2) Deduction of Target Relations, and 3) Translation into Natural Language Prompts.

\subsubsection{Adjacency and Boolean Matrix Construction} 
    \begin{itemize}
        \item Family Tree Reasoning: We generate four $n \times n$ Boolean matrices to represent the basic family relations: \texttt{IsFather}$(x, y)$, \texttt{IsMother}$(x, y)$, \texttt{IsSon}$(x, y)$, and \texttt{IsDaughter}$(x, y)$. For instance, if the entry at $(i,j)$ in the father matrix is 1, it implies person $j$ is person $i$’s father. 
            
        \item General Graph Reasoning: We randomly generate a $n \times n$ adjacency matrix to represent a directed graph of $n$ nodes. If there is a directed edge from node $i$ to node $j$, the entry at $(i,j)$ is set to 1; otherwise, it is 0.
        
    \end{itemize}
    All generated data are verified to ensure logical consistency, particularly for family relationships. Our implementation is adapted from the benchmarks provided in~\cite{dong2019neural}.

    An illustrative example of the Boolean matrices for the family tree reasoning benchmark with $n=10$ is shown below:
    \begin{center}
        \begin{tcolorbox} [size=title, title=Example - Boolean Matrices for Family Tree, breakable]
    Father matrix: 
    \[
    \begin{bmatrix}
      0& 0& 0& 0& 0& 0& 0& 0& 0& 0\\
      0& 0& 0& 0& 0& 0& 0& 0& 0& 0\\
      0& 0& 0& 0& 0& 0& 0& 0& 0& 0\\
      0& 0& 1& 0& 0& 0& 0& 0& 0& 0\\
      1& 0& 0& 0& 0& 0& 0& 0& 0& 0\\
      0& 0& 1& 0& 0& 0& 0& 0& 0& 0\\
      0& 0& 0& 0& 0& 0& 0& 0& 0& 0\\
      0& 0& 1& 0& 0& 0& 0& 0& 0& 0\\
      0& 0& 1& 0& 0& 0& 0& 0& 0& 0\\
      1& 0& 0& 0& 0& 0& 0& 0& 0& 0\\
    \end{bmatrix}
    \] 

    Mother matrix:
    \[
    \begin{bmatrix}
      0& 0& 0& 0& 0& 0& 0& 0& 0& 0\\
      0& 0& 0& 0& 0& 0& 0& 0& 0& 0\\
      0& 0& 0& 0& 0& 0& 0& 0& 0& 0\\
      0& 0& 0& 0& 0& 0& 0& 0& 0& 1\\
      0& 1& 0& 0& 0& 0& 0& 0& 0& 0\\
      0& 0& 0& 0& 0& 0& 0& 0& 0& 1\\
      0& 0& 0& 0& 0& 0& 0& 0& 0& 0\\
      0& 0& 0& 0& 0& 0& 0& 0& 0& 1\\
      0& 0& 0& 0& 0& 0& 0& 0& 0& 1\\
      0& 1& 0& 0& 0& 0& 0& 0& 0& 0\\
    \end{bmatrix}
    \] 

    Son matrix:
    \[
    \begin{bmatrix}
      0& 0& 0& 0& 1& 0& 0& 0& 0& 0\\
      0& 0& 0& 0& 1& 0& 0& 0& 0& 0\\
      0& 0& 0& 1& 0& 0& 0& 1& 1& 0\\
      0& 0& 0& 0& 0& 0& 0& 0& 0& 0\\
      0& 0& 0& 0& 0& 0& 0& 0& 0& 0\\
      0& 0& 0& 0& 0& 0& 0& 0& 0& 0\\
      0& 0& 0& 0& 0& 0& 0& 0& 0& 0\\
      0& 0& 0& 0& 0& 0& 0& 0& 0& 0\\
      0& 0& 0& 0& 0& 0& 0& 0& 0& 0\\
      0& 0& 0& 1& 0& 0& 0& 1& 1& 0\\
    \end{bmatrix}
    \]    
    
    Daughter matrix:
    \[
    \begin{bmatrix}
      0& 0& 0& 0& 0& 0& 0& 0& 0& 1\\
      0& 0& 0& 0& 0& 0& 0& 0& 0& 1\\
      0& 0& 0& 0& 0& 1& 0& 0& 0& 0\\
      0& 0& 0& 0& 0& 0& 0& 0& 0& 0\\
      0& 0& 0& 0& 0& 0& 0& 0& 0& 0\\
      0& 0& 0& 0& 0& 0& 0& 0& 0& 0\\
      0& 0& 0& 0& 0& 0& 0& 0& 0& 0\\
      0& 0& 0& 0& 0& 0& 0& 0& 0& 0\\
      0& 0& 0& 0& 0& 0& 0& 0& 0& 0\\
      0& 0& 0& 0& 0& 1& 0& 0& 0& 0\\
    \end{bmatrix}
    \]     
    \end{tcolorbox}
    \end{center}   

\subsubsection{Deduction of Target Relations}
    We apply the relational reasoning logic described in Section~\ref{sec:3} to deduce higher-order relations such as \texttt{HasSister}$(x)$ and \texttt{IsGrandson}$(x, y)$ etc. The output is expressed in Boolean vectors or matrices. On top of our illustrative example of father, mother, daughter and son matrices above, the Boolean matrix representing the \texttt{IsGrandson}$(x,y)$ relation is:

    \begin{center}
    \begin{tcolorbox} [size=title, title= Example - Grandson Relation, breakable]
    Grandson matrix:
    \[
    \begin{bmatrix}
       0& 0& 0& 1& 0& 0& 0& 1& 1& 0\\
       0& 0& 0& 1& 0& 0& 0& 1& 1& 0\\
       0& 0& 0& 0& 0& 0& 0& 0& 0& 0\\
       0& 0& 0& 0& 0& 0& 0& 0& 0& 0\\
       0& 0& 0& 0& 0& 0& 0& 0& 0& 0\\
       0& 0& 0& 0& 0& 0& 0& 0& 0& 0\\
       0& 0& 0& 0& 0& 0& 0& 0& 0& 0\\
       0& 0& 0& 0& 0& 0& 0& 0& 0& 0\\
       0& 0& 0& 0& 0& 0& 0& 0& 0& 0\\
       0& 0& 0& 0& 0& 0& 0& 0& 0& 0\\
    \end{bmatrix}
    \]  
    \end{tcolorbox}
    \end{center}

    \subsubsection{Translation into Natural Language Prompts}

    The matrix-based data are then converted into natural language for prompting the LLMs. Each prompt consists of:
    \begin{itemize}
        \item A system prompt that defines the LLM's task and expected response format (JSON).
    
        \item A user prompt describing the base relations (from the generated matrices) in natural language. The facts are shuffled to prevent positional biases.
    \end{itemize}
    
    Continuing with our illustrative example of \texttt{IsGrandson}$(x,y)$ with $n = 10$, the corresponding system prompt and user prompt are:

    \begin{center}
        \begin{tcolorbox} [size=title, title=Example - System Prompt (Family Tree), breakable]
    The user will provide a question performing reasoning on a family tree.
    
    Please give some reasoning steps briefly and give your matrix in a list of list in JSON format. 

~

    EXAMPLE JSON OUTPUT:
    
    \{
    
        ``Brief Reasoning Steps'': ``Some brief reasoning steps.'',
        
        ``Matrix'': [[1, 2, 3], [-1, 2, 3], [1, -1, 2]]
        
    \}
        \end{tcolorbox}
    \end{center}
    
    \begin{center}
        \begin{tcolorbox} [size=title, title=Example - User Prompt (Family Tree), breakable]
        You are an agent who determines the relations in a family.     
        
        For a family tree containing 10 family members, which is depicted with 4 kinds of relations: father, mother, son, daughter.     

        ~
        
        The relations are: 
        P1 is P9's mother. 
        P2 is P7's father. 
        P9 is P5's mother. 
        P3 is P9's son. 
        P9 is P7's mother. 
        P7 is P2's son. 
        P0 is P4's father. 
        P4 is P1's son. 
        P9 is P8's mother. 
        P9 is P3's mother.
        P5 is P9's daughter. 
        P5 is P2's daughter. 
        P9 is P1's daughter. 
        P3 is P2's son. 
        P1 is P4's mother. 
        P2 is P5's father. 
        P9 is P0's daughter. 
        P7 is P9's son. 
        P4 is P0's son. 
        P0 is P9's father. 
        P8 is P2's son. 
        P8 is P9's son. 
        P2 is P3's father. 
        P2 is P8's father. 

        ~
        
        Now, from the above-given facts, you have to determine the relation grandsons.  
        
        You must give the reasoning process and you must give the final answer in a 10-by-10 matrix, where the i,j-th entry is 1 if Pj is Pi's grandson, 0 otherwise. 

        ~
        
        Question: who has the relation grandson?     
        
        I need the matrix for further processing. 
        
        Do not include anything other than the matrix in python-readable format in ``Matrix''. 
        
        Your answer shall be in JSON format.     

        ~
        
        Answer: 
        \end{tcolorbox}
    \end{center}

Details of all the generation processes of Adjacency and Boolean matrices, deduction of target relations and the formulations of system prompts and user prompts can be found in our code repository. 

\subsection{Experimental Setup}

We generate datasets of size 100 for each of the six benchmark problems across three different values of $n$: 10, 20, and 40. This yields 18 distinct evaluation sets. All generated data are with non-trivial target relations, i.e., the Boolean matrices of target relations are not zero matrices. Each prompt is input to DeepSeek-R1, DeepSeek-V3, and GPT-4o in a zero-shot approach, i.e., without any in-context examples. The models' JSON responses are parsed, and the output matrices are compared to ground truth for evaluation.

\subsection{Evaluation Metrics}

To quantitatively assess model performance, we use F1-score as the primary metric due to its effectiveness in measuring classification accuracy, especially under class imbalance.

\begin{itemize}
    \item \textbf{Binary Classification Tasks}: 
    For problems such as \texttt{HasSister}$(x)$, \texttt{IsGrandson}$(x,y)$, \texttt{IsAunt}$(x,y)$, \texttt{IsPaternalGreatAunt}$(x,y)$ and \texttt{Connectivity}$(x,y)$, we compute the standard F1-score:
    \[
    \text{Standard F1-score} = \frac{2  \times \text{Precision}\times\text{Recall}}{\text{Precision}+\text{Recall}}.
    \]

    \item \textbf{Multiclass Classification Task}:
    For \texttt{Shortest}$(x,y)$, we adopt Macro-averaged F1-score to ensure balanced evaluation across all classes, regardless of their individual frequencies.

    \item \textbf{Special Evaluation Cases}:
    To maintain evaluation consistency, we define the following conditions:

    \begin{enumerate}
        \item If a model fails to return a valid JSON format, an F1-score of 0 is assigned.
    
        \item If the returned matrix has an incorrect shape (i.e., not $n \times n$), an F1-score of 0 is also assigned.
    \end{enumerate} 
\end{itemize}

\begin{table*}[!htbp]
\caption{Average F1-scores of LLMs on Family Tree and General Graph Reasoning Benchmark Problems over 100 Runs}
\centering
  \resizebox{\textwidth}{!}{
\begin{tblr}{
  cells = {c},
  cell{1}{1} = {r=2}{},
  cell{1}{2} = {c=3}{},
  cell{1}{5} = {c=3}{},
  cell{1}{8} = {c=3}{},
  vline{5,8} = {2-11}{},
  hline{1,9} = {-}{0.2em},
  hline{7} = {1-13}{},
  hline{2} = {1-13}{},
  hline{3} = {-}{},
}
Benchmark Problem          & $n=10$ &  &    &  $n=20$ & &   &  $n=40$ &   &  \\
               & GPT-4o        & DeepSeek-V3    & DeepSeek-R1    & GPT-4o        & DeepSeek-V3    & DeepSeek-R1    & GPT-4o        & DeepSeek-V3    & DeepSeek-R1 \\
\texttt{HasSister}$(x)$ & 0.516  &  0.542  &  \textbf{0.803}  &  0.599  & 0.529 & \textbf{0.975} & 0.667 & 0.590 &  \textbf{0.893} \\
\texttt{IsGrandson}$(x,y)$  & 0.070  &  0.180  & \textbf{0.778} & 0.000   & 0.003 & \textbf{0.921} & 0.000 & 0.000 &  0.000 \\
\texttt{IsAunt}$(x,y)$ &  0.017  &  0.197  & \textbf{0.980} &  0.000  & 0.003 & \textbf{0.832} & 0.000 & 0.000 &  0.000 \\
\texttt{IsPaternalGreatAunt}$(x,y)$ & 0.000 &  0.065 & \textbf{0.390}  & 0.000 & 0.000  &  \textbf{0.367} & 0.000 & 0.000 &  0.000 \\
\texttt{Connectivity}$(x,y)$ & 0.400   &  0.561  & \textbf{0.743} & 0.161 & 0.153 & \textbf{0.432} & 0.000 & 0.070 & \textbf{0.103} \\
\texttt{Shortest}$(x,y)$  & 0.466 &  0.491 & \textbf{0.969} & 0.060 & 0.243 & \textbf{0.446} & 0.000 & \textbf{0.238} &  0.201 \\
\end{tblr}
}
\label{tab:results}
\vspace{-1em}
\end{table*}

\section{Result Analysis and Discussion}\label{sec:5}

In this section, we analyze the experimental results, examine their implications in depth, and outline future directions. The average F1-scores of the LLMs across six family tree and general graph reasoning benchmark problems, evaluated over 100 runs, are summarized in Table~\ref{tab:results}.

\subsection{Family Tree Reasoning}

Across all values of $n=10$, $20$ and $40$, DeepSeek-R1 consistently achieves the highest average F1-scores, except at $n=40$ where all three evaluated models fail completely on the problems \texttt{IsGrandson}$(x,y)$, \texttt{IsAunt}$(x,y)$ and \texttt{IsPaternalGreatAunt}$(x,y)$, each yielding an F1-score of 0.000. Also, for problems such as \texttt{HasSister}$(x)$, \texttt{IsGrandson}$(x,y)$ and \texttt{IsAunt}$(x,y)$, DeepSeek--R1 reaches impressive average F1-scores as high as 0.9 at both $n=10$ and $20$, providing strong evidence of its advanced capability in deep relational reasoning.

However, performance of DeepSeek-R1 on \texttt{IsPaternalGreatAunt}$(x,y)$ remains relatively modest even at smaller scales ($n=10$ and 20), with F1-scores only slightly above 0.3. This likely reflects the higher logical complexity involved in deducing the paternal great aunt relation compared to other family relations.

Among all problems, \texttt{HasSister}$(x)$ is the least challenging, which explains why all three models achieve their highest scores on this problem (all above 0.5). This further supports the notion that DeepSeek-R1's performance edge becomes more pronounced as relational complexity increases.

The performances of GPT-4o and DeepSeek-V3 are generally comparable across all problems. While DeepSeek-V3 slightly outperforms GPT-4o in certain cases, the differences are not statistically significant. As both models lack specialized deep reasoning mechanisms, these findings indirectly affirm the superiority of DeepSeek-R1 in family tree reasoning problems.

\subsection{General Graph Reasoning}

Similar trends are observed in the general graph reasoning problems. At $n=10$ and 20, DeepSeek-R1 consistently outperforms both GPT-4o and DeepSeek-V3 on texttt{Connectivity}$(x,y)$ and \texttt{Shortest}$(x,y)$, achieving average F1-scores above 0.7 at $n=10$ and above 0.4 at $n=20$.

In contrast, both GPT-4o and DeepSeek-V3 demonstrate only moderate performance, with DeepSeek-V3 showing a slight edge in certain runs. However, when $n=40$, all three LLMs experience significant performance degradation, with average F1-scores dropping below 0.2. In some cases, such as with GPT-4o, the F1-scores even fall to zero, reflecting a complete failure. This mirrors the models’ struggles with more complex general graph reasoning problems, highlighting a shared limitation when scaling to deeper relational depths.

Taken together, these results highlight DeepSeek-R1 as the most competent LLM for both types of relational reasoning, offering a marked performance advantage over models lacking deep reasoning capabilities.

\subsection{In-depth Discussion}
\paragraph{Intrinsic Token Limits} 
All LLMs inherently face token limits dictated by their architecture. This constraint appears to be a key factor behind DeepSeek-R1’s failure at $n=40$ across most problems except \texttt{HasSister}$(x)$. Closer inspection of failed runs reveals instances where the reasoning process is truncated, and the output matrix is either incomplete or malformed. In our evaluation protocol, any such output is assigned an F1-score of 0, which penalizes performance in evaluation.

The same limitation likely contributes to the performance breakdowns of GPT-4o and DeepSeek-V3 at higher $n$ values. As prompt length and complexity grow with $n$, these models are unable to complete their reasoning process and reply within the allowed token budget.

\paragraph{Long Chain-of-Thought}
DeepSeek-R1’s exceptional performance can be attributed in part to its long Chain-of-Thought (CoT) reasoning, a core component of its architecture. Careful examination of its output reveals rich, structured reasoning not found in GPT-4o or DeepSeek-V3 responses.

Below, we present an illustrative excerpt from DeepSeek-R1’s response to an \texttt{IsGrandson}$(x,y)$ problem (described in Section~\ref{sec4:setup}). For clarity, we highlight several distinct reasoning behaviors using color-coded annotations:

\begin{center}
    \begin{tcolorbox} [size=title, title=Example - Deep Reasoning Content of DeepSeek-R1, breakable]
        
    DeepSeek-R1: $<$think$>$ 
    
    Okay, let's try to figure out who the grandsons are in this family tree. First, I need to parse all the given relations and build a structure to represent the family members and their connections. Let's start by listing out all the given facts one by one and see how they connect.

~
    \color{red}
    
    The family members are P0 to P9. The relations given are:
    
    1. P4 is P1's son.
    
    2. P2 is P5's father.

    ~
    
    \color{red}3. P1 is P9's mother.

    $<$Omitted by the authors$>$... 

    24. P1 is P4's mother.

    ~

    \color{blue}

    Hmm, some of these might be redundant or reinforcing existing relationships. Let me start by mapping out each \color{blue}person's parents and children.

    ~

    Starting with P0. From fact 11, P0 is P4's father. Fact 18 says P0 is P9's father. Fact 17 says P9 is P0's daughter. So P0 has children P4 (son) and P9 (daughter).

    ~

    P1: Fact 3 says P1 is P9's mother. Fact 12 clarifies that P9 is P1's daughter. Fact 24 says P1 is P4's mother. So P1's children are P4 (son) and P9 (daughter). But wait, fact 1 says P4 is P1's son, which aligns with that. Also, \color{blue}P9 is their daughter.

~

    ...$<$Content omitted for more compact presentation$>$...

~

    P9: Her parents are P0 (father) and P1 (mother) from facts 17,18 and 3,12. Her children are P3 (son), P5 (daughter?), wait no. Let's check. Fact 7 says P3 is P9's son. Fact 8: P8 is P9's son. Fact 15: P5 is P9's daughter. Fact 20: P7 is P9's son. Wait, that's conflicting. Let me check again.

    \color{black}

    ~

    \color{brown}

    Wait, fact 6: P9 is P3's mother. Fact 7: P3 is P9's son. So P3 is a child of P9. Fact 8: P8 is P9's son. Fact 13: P9 is P5's mother. Fact 15: P5 is P9's daughter. Fact 16: P9 is P7's mother. Fact 20: P7 is P9's son. Fact 22: P9 is P8's mother. So P9's children are P3 (son), P5 (daughter), P7 (son), P8 (son). Wait, that's a lot. So P9 has four children: P3, P5, P7, P8. But how? Because P9's parents are P0 and P1. And P9's spouse must be P2, since P2 is the father \color{brown}of P3, P5, P7, P8. Because P3's father is P2 (fact 5,19), P5's father is P2 (fact 2,21), P7's father is P2 (10,14), P8's father is P2 (4,23). So P2 and P9 are spouses, parents of these four children.

    ~

    \color{brown}
    So P2 and P9 are a couple, with children P3, P5, P7, P8. P3 is male, P5 female, P7 male, P8 male.

    ~

    Now, P0 and P1 are parents of P4 (son) and P9 (daughter). So P9 is the daughter of P0 and P1, and married to P2, having four children. P2's other children? Wait, no, P2's children are with P9. Wait, but P2's children are P3, P5, P7, P8 (all with P9), and possibly others? Let me check the facts again.

    ~

    \color{brown}Wait, the given facts mention that P2 is the father of P5, P3, P7, P8. All these have P9 as their mother. So P2 and P9 are the parents.


    ~

    So the family structure so far:

    - P0 (male) and P1 (female) have children P4 (male) and \color{brown}P9 (female).

    - P9 (female) and P2 (male) are married and have children: P3 (male), P5 (female), P7 (male), P8 (male).

    ~

    Additionally, let's check if there are other parents or children mentioned.

    Looking at fact 24: P1 is P4's mother. Which we already have.

    ~

    \color{violet}

    So now, we need to find grandsons. A grandson of Pi is a son of Pi's child. So for each person Pi, we need to look at their children, then check if those children have sons, and those sons would be Pi's grandsons.

    ~

    So the steps would be:

    1. For each Pi, find their children (sons and daughters).
    
    2. For each child of Pi, check if that child has any sons (because a grandson is a son of one's child).
    
    ~
    
    So for each Pi, collect all sons of their children (both sons and daughters, because a daughter's son is also a grandson).

    ~

    Wait, the question is to determine grandsons. So the grandson is the son of one's child, regardless of the gender of the child. So if Pi has a daughter who has a son, that son is Pi's grandson.

    \color{black}

    ~

    \color{teal}

    Now, let's check each person:

     ~

    Starting with P0:

    Children: P4 (son) and P9 (daughter).

    P4's children: Are there any? Looking at the given facts, P4's relations: he is the son of P0 and P1. Are there \color{teal}any mentions of P4 having children? Let me check all facts again. The facts given don't mention P4 having any children. So P4 has no sons mentioned. So P0's grandsons would be the sons of P9. P9's sons are P3, P7, P8. So those three are grandsons of P0.

     ~

    \color{teal}

    Next, P1:

    Children: P4 and P9. Again, P4 has no children mentioned, so P1's grandsons would be P9's sons: P3, P7, P8.

    ~
    
    ...$<$Content omitted for more compact presentation$>$...

    ~
    
    P9: 
    
    Her children are P3, P5, P7, P8. None of them have children mentioned. So P9 has no grandsons.

    ~

    \color{teal} Wait, but wait. Let's confirm again. The facts given only mention the relations provided. If a person isn't mentioned as a parent, we assume they don't have children in this family tree. So unless stated, we can't assume any.

    ~
    
    \color{teal}Therefore, for each Pi, their grandsons are the sons of their children, but only if those children have sons as per the given data.
    
    So let's go through each Pi:

    ~
    
    P0: Children P4 and P9.
    
    - P4 has no children mentioned.
    
    - P9 has sons P3, P7, P8.
    
    So P0's grandsons: P3, P7, P8.

    ~
    
    P1: Children P4 and P9.
    
    Same as P0, grandsons P3, P7, P8.

~
    \color{black}

    ...$<$Content omitted for more compact presentation$>$...
    
    \end{tcolorbox}
    \end{center}

\begin{enumerate}

    \item  \textbf{Summarization and Comprehension} (\textcolor{red}{Red}): DeepSeek-R1 effectively summarizes the relationships in the user prompt, listing them in a clear and coherent manner. This demonstrates foundational capabilities in information extraction and synthesis.

    \item \textbf{Abstraction and Filtering} (\textcolor{blue}{Blue}): DeepSeek-R1 identifies potentially redundant or conflicting information and cross-examines each individual. It also deduces intermediate relations, such as parent and child, before addressing the target grandson relation. This reflects structured reasoning with abstraction.

    \item \textbf{Verification and Confirmation} (\textcolor{brown}{Brown}, \textcolor{teal}{Teal}): DeepSeek-R1 repeatedly verifies and validates intermediate deductions. While this process may appear somewhat unstructured, it mimics a natural human reasoning approach, and contributes to robust inference. For example, DeepSeek-R1 verifies and confirms its findings on parent and child relations repeatedly before proceeding to deduce the grandson relation in the part highlighted in \textcolor{brown}{brown} color. Also, it evaluates each person’s potential grandsons, repeating the verification process multiple times before finalizing its conclusions in the part highlighted in \textcolor{teal}{teal} color.

    \item \textbf{Planning} (\textcolor{violet}{Violet}): DeepSeek-R1 articulates a plan to deduce the grandson relation, though interestingly, this plan emerges midway through the reasoning process rather than at the outset. This raises intriguing questions about the internal reasoning loop and how DeepSeek-R1 dynamically constructs strategies during inference.

    In detail, this phenomenon demonstrates the competent planning ability inside the deep reasoning capabilities of DeepSeek-R1, a particularly important ability for handling complicated reasoning problems which can hardly be solved in a short period of time and reasoning process. However, this raises the question why such plan of checking grandson relation suddenly pops up in the middle of the reasoning process rather than in the beginning of the reasoning process. In particular, it draws ones to ask how DeepSeek-R1 knows that it should deduce intermediate relations like parent and child before addressing the relation grandson if such plan is only derived afterwards. While DeepSeek-R1 demonstrates strong planning ability, the reasoning process may not be entirely well-organized and structured, suggesting that, despite appearing coherent, the underlying reasoning rationale may not always be fully sound and deserve further investigation.
    
\end{enumerate}

Overall, DeepSeek-R1 demonstrates a unique blend of planning, structured reasoning, iterative validation, and robust abstraction, all of which are hallmarks of advanced deep reasoning. However, due to token limitations, this process becomes infeasible at $n=40$, where reasoning depth and prompt length scale significantly.

Despite architectural limitations imposed by token constraints, DeepSeek-R1 consistently demonstrates superior performance in deep relational reasoning across both family tree and general graph reasoning benchmarks. Its long CoT paradigm enables it to engage in rich, human-like reasoning processes that remain unmatched by current models without dedicated deep reasoning mechanisms. These findings highlight both the promise and limitations of current LLMs in advancing toward more sophisticated cognitive capabilities.

\subsection{Future Directions}

\paragraph{Multimodal Reasoning} 
Our findings reaffirm that deep reasoning capabilities are critical for tackling relational reasoning tasks. However, recent studies have shown that LLMs may exhibit greater proficiency in multimodal reasoning in domains such as time-series analysis than in purely text-based reasoning tasks~\cite{zhou2024can}. Multimodal reasoning involves processing and integrating information from non-textual sources such as images or videos, enabling spatial, geometric, and visual cognition. Given that humans naturally rely on spatial and visual reasoning for a wide range of everyday tasks~\cite{ge2024worldgpt,wu2024next}, an important future direction is to explore whether relational reasoning in LLMs can be further enhanced through multimodal prompts. This raises the question: could visual or diagrammatic representations of relational structures, for example, family trees or graphs, serve as a more effective input modality for enabling and strengthening deep relational reasoning?

\paragraph{Deeper Investigation into Long Chain-of-Thought Processes}

As illustrated in our analysis of DeepSee--R1’s reasoning trace, LLMs equipped with deep reasoning capabilities can engage in remarkably thorough and robust inference prior to generating a final answer. However, as noted in the planning phase of the reasoning process, DeepSeek-R1 occasionally constructs its plan midway through reasoning rather than at the outset. This observation raises important concerns regarding the coherence and soundness of the internal reasoning rationale. While the output may appear logically consistent, the underlying reasoning process may, at times, lack proper structure or contain subtle flaws.

This opens up a rich line of inquiry into the internal reasoning dynamics of such models. A promising future direction involves conducting fine-grained analysis of reasoning traces that ultimately lead to incorrect conclusions. By scrutinizing these examples, we may uncover latent error patterns, such as premature assumptions, flawed inferences, or inconsistencies in intermediate steps, that can inform the design of novel training objectives, architectural modifications, or reasoning frameworks. Such efforts could significantly enhance the robustness and interpretability of deep reasoning in LLMs.

\section{Conclusion}\label{sec:6}

In this paper, we systematically evaluated the deep relational reasoning capabilities of state-of-the-art LLMs through a series of benchmark problems in family tree and general graph reasoning. Among the three models evaluated, DeepSeek-R1, DeepSeek-V3 and GPT-4o, our experimental results consistently demonstrate the superior performance of DeepSeek-R1, especially at smaller problem scales ($n = 10$ and 20). Its ability to perform accurate and structured logical deductions across multiple layers of relational dependencies confirms its potential for deep reasoning tasks.

However, our analysis also reveals key limitations. All models, including DeepSeek-R1, exhibit performance degradation at larger problem sizes ($n = 40$), primarily due to intrinsic token limits and incomplete output generations. A detailed investigation into DeepSeek-R1’s reasoning traces shows that, while its long Chain-of-Thought processes are a key strength, they are not always well-structured or logically sound throughout, raising important questions about the underlying reasoning mechanisms of such models.

Looking ahead, we identify two promising directions for future work. First, the potential of multimodal reasoning, incorporating visual and spatial inputs, may further enhance the relational reasoning capabilities of LLMs. Second, a deeper examination of long Chain-of-Thought responses, especially those that lead to incorrect outcomes, can offer insights into the cognitive patterns of LLMs and guide improvements in both training and inference strategies.

Overall, our study highlights both the promise and current limitations of LLMs in performing deep relational reasoning. It provides an empirical and analytical foundation for future research aimed at building more robust, interpretable, and human-aligned reasoning systems.

\bibliographystyle{IEEEtran}
\bibliography{cas-refs}

\end{document}